\documentclass[final,3p,times]{elsarticle}

\usepackage{geometry}
\usepackage{graphicx}
\usepackage{amssymb}
\usepackage{booktabs}

\usepackage{caption}
\usepackage{subcaption}
\usepackage{lineno}
\usepackage{array}
\usepackage{float}
\usepackage{tcolorbox}
\usepackage{amsmath,amsthm}
\usepackage{mathrsfs}
\usepackage[colorlinks=true, allcolors=blue]{hyperref}
\usepackage[nodots]{numcompress}
\usepackage{arydshln}

\usepackage{algorithm2e}

\usepackage{multirow}

\biboptions{sort&compress}
\graphicspath{{Figures/}}

\theoremstyle{remark}

\journal{}

\begin{document}
\begin{frontmatter}

\title{One-Step Flow Matching for Generative Modeling of Path-Dependent Physical Fields}

\author[label1]{Yijing Zhou}
\author[label1,label2]{Jasmin Jelovica\corref{cor1}}
\ead{jjelovica@mech.ubc.ca}
\date{}
\address[label1]{Department of Mechanical Engineering, The University of British Columbia, Vancouver, Canada}
\address[label2]{Department of Civil Engineering, The University of British Columbia, Vancouver, Canada}

\cortext[cor1]{Corresponding author}

\begin{abstract}

Physical simulations for intricate geometries with path-dependent constitutive models face difficulties due to the enormous computational cost they require. Recently, the emergence of generative AI models, which succeed in image and video synthesis tasks, has provided a promise to further improve simulations. Although U-Net-based denoising diffusion probabilistic models (DDPMs) have been adopted for elastic stress field generation, they typically require hundreds of sampling steps, and applications of generative models to path-dependent, e.g. plastic, stress fields remain very limited. In this work, we propose a novel flow matching (FM) model based on a transformer backbone for high-resolution path-dependent stress field generation with stochastic loading-unloading paths and geometry. The proposed model operates within the latent space of a variational autoencoder (VAE) and formulates the simulation of plastic fields as a video synthesis task, directly generating the stress fields across all time steps. Meanwhile, we design a non-Gaussian source distribution for flow matching, such that crossings among conditional transport paths are reduced during training. This enables our model to generate satisfactory samples in one step without relying on distillation. In addition, we introduce token-level loading embeddings and two auxiliary networks to further enhance the model performance in path-dependent simulation. The results demonstrate that, even with a limited training dataset, our model can accurately generate high-resolution path-dependent fields. It is much more computationally efficient than finite element analysis, providing a speedup of 6 to 7 times over FEM on CPUs and approximately two orders of magnitude speedup on consumer-grade GPUs.

\end{abstract}


\end{frontmatter}



\section{Introduction}\label{sec:Introduction}

Fast and reliable simulation methods are of critical importance for modern mechanical, aerospace, and ocean engineering. Traditional numerical methods, such as the finite element method (FEM), have long been the predominant approach for performing simulations and have made substantial contributions to the advancement of engineering. However, when traditional numerical methods are applied to complex geometries and sophisticated constitutive models, very fine mesh densities and numerous iterations are required to achieve satisfactory results, which can substantially increase the computational cost and, in some cases, even render the simulation infeasible. Thus, new methods are needed to address those challenges. Fortunately, the emergence of machine learning (ML) methods has provided a new perspective for further accelerating simulations. A number of review articles \cite{bishara2023state, bock2019review, LIU2021109152, fish2021mesoscopic, liu2022eighty} summarize the latest advances in machine learning for computational mechanics.

Sequence modeling architectures have been widely demonstrated in the literature to play an important role in constitutive modeling and simulations. Recurrent neural networks (RNNs), along with their gated variants, including gated recurrent units (GRUs) and long short-term memory (LSTM) networks, have been used in computational mechanics for a considerable period of time. RNNs have been used to learn inelastic constitutive models for stress prediction \cite{mozaffar2019deep, bonatti2022importance, zhang2021application, im2021surrogate, heidenreich2024recurrent, guan2023neural, tancogne2021recurrent, dettmer2024framework, chen2021recurrent} in some researchers' work. Other investigators applied RNN to learn the path-dependent constitutive relationships in homogenization-based multiscale methods \cite{zhou2025machine, WU2020113234, ghavamian2019accelerating, logarzo2021smart, wang2018multiscale, haghighi2022single, calleja2023development, wu2024self, kim2023surrogate, vijayaraghavan2023data}. RNN models can also be used for surrogate modeling of foam \cite{gorji2020potential} and short-fiber composite materials \cite{friemann2023micromechanics}. Moreover, some works on physical information combined RNN models \cite{tandale2022physics, maia2023physically, deng2024data, borkowski2022recurrent, qu2021towards, maia2024physically} have been done, which give us a new aspect. Furthermore, RNNs have been utilized on hybrid data-physics driven reduced-order homogenization \cite{fish2023data}, transfer learning-based works \cite{heidenreich2024transfer, qu2023data}, and comparative study \cite{borkowski2022recurrent, tandale2023recurrent}. However, RNNs often struggle with long-sequence modeling, whereas the success of transformers \cite{vaswani2017attention} in natural language processing (NLP) has drawn considerable attention to their powerful ability to capture long-range dependencies. Consequently, researchers have begun to explore transformer-based models for constitutive modeling \cite{zhou2025vit, zhongbo2024pre}, field prediction \cite{buehler2022fieldperceiver,buehler2022end}, and material design \cite{yang2021words}.

Convolutional neural networks (CNNs) and graph neural networks (GNNs) have been employed for processing image data and graph-structured data, respectively. CNNs are a type of fundamental architecture for computer vision tasks, which have multiple applications in computational mechanics. CNNs in some works have been utilized to map the spatial arrangement of fibers to the corresponding stress fields \cite{BHADURI2022109879, gupta2023accelerated}, which accelerates the simulation of composite materials. This approach has been expanded to elastoplastic \cite{saha2024prediction} and elasto-viscoplastic \cite{khorrami2023artificial} constitutive modeling of composite materials. CNNs are also used to predict the properties of composite \cite{yang2019using, su2023three, chang2022predicting, mahdi2024lattice} or nanoporous \cite{mianroodi2022lossless} materials. Furthermore, CNNs can be utilized for the stress-strain curves prediction of composites along the failure path \cite{yang2020prediction}, anisotropic effective material properties prediction of random inclusions RVE \cite{rao2020three}, computational homogenization \cite{peng2022ph}, and topology optimization \cite{yan2025kato, yan6184689katosuper}. Furthermore, the applications of GNNs have also emerged in recent research. GNNs are utilized for numerical simulation acceleration \cite{gulakala2023graph, jiang2023graph, maurizi2022predicting, cai2024efficient, cai2026heterogeneous, cai2026hybrid}, material properties prediction \cite{dai2021graph}, material fracture mechanics behavior prediction \cite{perera2022graph, karapiperis2023prediction}, and material fatigue detection \cite{thomas2023materials}.

Recently, the rise and success of generative artificial intelligence (GenAI) have attracted increasing interest from the engineering community. Two of the most representative classes of models in generative AI are denoising diffusion probabilistic models (DDPMs) \cite{NEURIPS2020_4c5bcfec, song2021scorebased}, which learn noise patterns, and flow matching (FM) \cite{lipman2023flow, lipman2024flow, JMLR:v26:23-1605} models, which learn time-dependent vector fields in sample space. In engineering, research on generative AI is still in its early stages, with existing studies focusing on simulation for elastic or hyperelastic stress \cite{jadhav2023stressd, kota2026hybrid} and fluid fields \cite{shu2023physics, KASHEFI2026119037}. Research on flow matching in solid mechanics is very scarce.

Several limitations in existing studies remain to be addressed. First, current approaches are generally unable to handle stress fields governed by path-dependent constitutive models, such as plasticity. However, plastic behavior with loading-unloading paths is highly prevalent and of significant importance in engineering applications. Second, diffusion models typically require hundreds of sampling steps to generate satisfactory samples, and even flow matching usually requires tens of steps, which is computationally expensive. Therefore, generative models capable of producing high-quality samples in only a few steps, or even a single step, could be of great importance to the engineering and AI community. Third, existing studies commonly adopt the U-Net as the backbone, whereas the diffusion transformer (DiT) backbone, which has become increasingly dominant in image and video synthesis, deserves further consideration because of its stronger capability in modeling time-dependent data. It should be noted that the term DiT was originally introduced because the authors of the original paper \cite{peebles2023scalable} employed it as the backbone of diffusion models. In the current AI community, this backbone is still commonly referred to as DiT, even with a flow matching training objective. For consistency, we follow this terminology; however, the model proposed in the current work is trained using flow matching rather than diffusion.

\begin{figure}[htp]
    \centering
    \includegraphics[width=\linewidth]{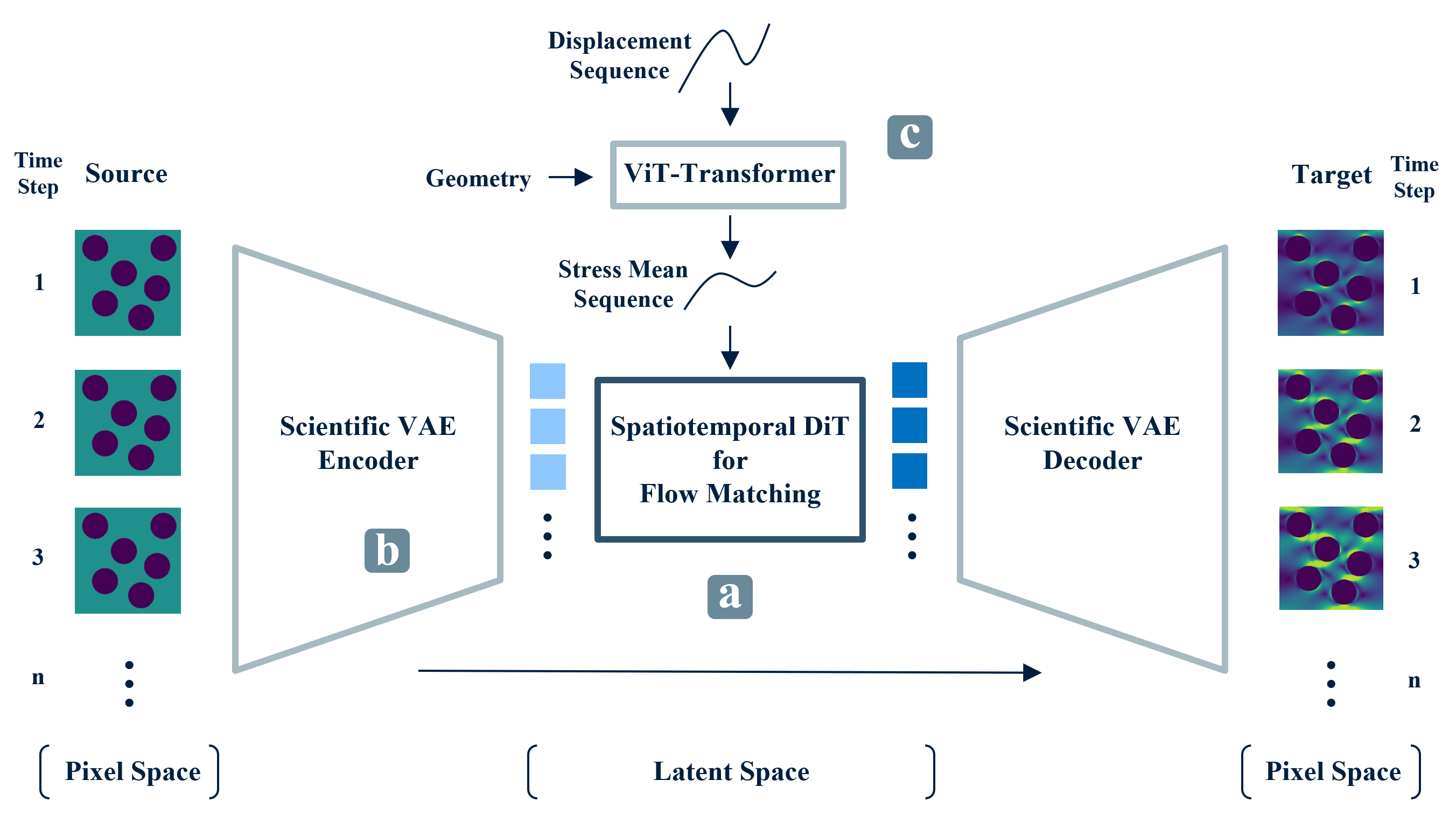}
    \caption{Overall schematic. (a) Spatiotemporal DiT backbone for learning the velocity field in the sample (latent) space; (b) Scientific VAE for dimensionality reduction; (c) ViT-Transformer for loading embedding preprocessing.}
    \label{fig:Graphical_Abstract}
\end{figure}

To overcome the challenges discussed above, this work proposes a novel flow matching model with a spatiotemporal DiT backbone to generate path-dependent von Mises stress fields from stochastic loading-unloading paths and random geometry. To further improve the efficiency and performance of the proposed framework, we pretrain two auxiliary networks, whose architectures, together with the main network, are illustrated in Figure \ref{fig:Graphical_Abstract}. This research makes the following main contributions:

\begin{itemize}
    \item A novel flow matching model with a geometry-based empirical distribution source is proposed for the high-resolution path-dependent stress field generation. The results demonstrate that it can synthesize satisfactory samples using only one sampling step or a very small number of steps.
    \item We propose a DiT backbone with token-level loading embedding, which can simultaneously learn the spatiotemporal dependencies across all frames while incorporating the effects of geometry, loading path, and constitutive model on the stress field. With this design, the flow matching model can directly generate stress fields over all time steps involving elastoplastic behavior.
    \item Two auxiliary networks, namely a scientific VAE and a ViT-Transformer, are introduced for dimensionality reduction and loading injection preprocessing, respectively. These auxiliary components substantially improve the efficiency and performance of the main network.
\end{itemize}

The structure of this paper is as follows. Section \ref{sec:Methodology} presents the methodology and provides a detailed introduction to flow matching, DiT, and the auxiliary networks. Section \ref{sec:Datasets} describes the design and generation procedure of the dataset. Section \ref{sec:Results} presents the model performance, computational efficiency, and generated samples in detail, followed by a discussion of the results. Section \ref{sec:Conclusion} summarizes the main findings.

\section{Methodology}\label{sec:Methodology}

\subsection{Flow Matching}\

Flow matching (FM) \cite{lipman2023flow} learns a time-dependent velocity field over the sample space, whose associated flow continuously transports probability mass from a source distribution to the target data distribution, thereby providing a simulation-free training framework for continuous normalizing flows (CNFs). FM enables the training of CNFs using alternative, non-diffusion probability paths. A particularly important approach is the use of Optimal Transport (OT) displacement interpolation to construct conditional probability paths. Compared with diffusion paths, these paths are more efficient, allow faster training and sampling, and lead to improved generalization. Notably, researchers find that sampling trajectories induced by diffusion paths may overshoot the target sample \cite{lipman2023flow}, thereby requiring unnecessary backtracking, whereas OT paths are guaranteed to remain straight. This enables flow matching to require substantially fewer sampling steps than diffusion models.

However, achieving one-step generative modeling with flow matching has been highly challenging. The reason is that the conditional trajectories connecting different noise samples and data points can intersect extensively in the sample space, causing the model to learn an averaged velocity over multiple directions at the same location. As a result, the learned flow field becomes curved, making one-step generation difficult. Therefore, flow matching typically still requires tens of sampling steps for progressive refinement, although this is already substantially more efficient than diffusion models. In this section, we first present the formulation of flow matching and then discuss how one-step generation can be achieved.

\subsubsection{Formulation}\

We regard each data point, which in this work corresponds to a path-dependent stress field containing all frames, as a point in a high-dimensional data space. Let $\mathbb{R}^d$ denote the data space, let $p_0(x)=\mathcal{N}(x\mid 0,I)$ be the prior distribution, and let $q(x_1)$ denote the unknown data distribution. A CNF is defined through the ordinary differential equation
\begin{equation}
    \frac{d}{dt}\phi_t(x)=v_t(\phi_t(x);\theta), 
    \qquad \phi_0(x)=x,
    \label{eq:cnf_ode}
\end{equation}
where $v_t(\cdot;\theta)$ is a neural vector field and $\phi_t$ is a time-dependent map, called a flow. If the flow map $\phi_t$ pushes $p_0$ forward to a probability path $p_t$, i.e.,
\begin{equation}
    p_t = [\phi_t]_{\ast}p_0,
    \label{eq:push_forward}
\end{equation}
then solving Equation \ref{eq:cnf_ode} from $t=0$ to $t=1$ transforms samples from the prior into samples from the learned data distribution.

The ideal FM objective assumes that a target probability path $p_t(x)$ and its generating vector field $u_t(x)$ are known, and trains $v_t$ by
\begin{equation}
    \mathcal{L}_{\mathrm{FM}}(\theta)
    =
    \mathbb{E}_{t\sim \mathcal{U}[0,1],\,x\sim p_t}
    \left[
    \left\|v_t(x;\theta)-u_t(x)\right\|^2
    \right].
    \label{eq:fm_loss}
\end{equation}
However, the marginal vector field $u_t(x)$ is generally intractable. FM resolves this difficulty by constructing conditional probability paths indexed by data samples $x_1\sim q(x_1)$. Specifically, given conditional paths $p_t(x\mid x_1)$, the marginal path is defined as
\begin{equation}
    p_t(x)=\int p_t(x\mid x_1)q(x_1)\,dx_1,
    \label{eq:marginal_path}
\end{equation}
and the corresponding marginal vector field can be written as
\begin{equation}
    u_t(x)=
    \int u_t(x\mid x_1)
    \frac{p_t(x\mid x_1)q(x_1)}{p_t(x)}
    \,dx_1.
    \label{eq:marginal_vf}
\end{equation}
Although Equation \ref{eq:marginal_vf} is not directly computable in practice, this identity shows that the marginal vector field is the conditional expectation of the conditional vector field under the posterior distribution. Therefore, we can prove \cite{lipman2023flow} that the Conditional Flow Matching objective
\begin{equation}
    \mathcal{L}_{\mathrm{CFM}}(\theta)
    =
    \mathbb{E}_{t\sim \mathcal{U}[0,1],\,x_1\sim q,\,x\sim p_t(\cdot\mid x_1)}
    \left[
    \left\|v_t(x;\theta)-u_t(x\mid x_1)\right\|^2
    \right]
    \label{eq:cfm_loss}
\end{equation}
has the same gradient with respect to $\theta$ as Equation \ref{eq:fm_loss}, up to a constant independent of $\theta$, i.e. 
\begin{equation}
\nabla_{\theta} \mathcal{L}_{\mathrm{FM}}(\theta)=\nabla_{\theta} \mathcal{L}_{\mathrm{CFM}}(\theta)
\label{eq:fm_cfm}
\end{equation}
Therefore, the model can be trained using only tractable conditional vector fields.

In the commonly used Gaussian-to-data OT conditional path in Flow Matching, the conditional probability path is chosen as a Gaussian path whose mean and standard deviation change linearly in time:
\begin{equation}
    p_t(x\mid x_1)
    =
    \mathcal{N}\!\left(
    x \mid t x_1,\,
    \left(1-(1-\sigma_{\min})t\right)^2 I
    \right),
    \label{eq:ot_conditional_path}
\end{equation}
where $\sigma_{\min}>0$ is a small terminal standard deviation. Equivalently, this path is induced by the conditional flow map
\begin{equation}
    \psi_t(x_0)
    =
    \left(1-(1-\sigma_{\min})t\right)x_0 + t x_1,
    \qquad x_0\sim \mathcal{N}(0,I).
    \label{eq:ot_flow_map}
\end{equation}
The map in Equation \ref{eq:ot_flow_map} is the OT displacement interpolation between the initial Gaussian distribution and the terminal Gaussian distribution centered at $x_1$. Its associated conditional vector field is
\begin{equation}
    u_t(x\mid x_1)
    =
    \frac{x_1-(1-\sigma_{\min})x}
    {1-(1-\sigma_{\min})t}.
    \label{eq:ot_conditional_vf}
\end{equation}
Using the reparameterization $x=\psi_t(x_0)$, the practical training objective becomes
\begin{equation}
    \mathcal{L}_{\mathrm{CFM}}(\theta)
    =
    \mathbb{E}_{t\sim \mathcal{U}[0,1],\,x_1\sim q,\,x_0\sim \mathcal{N}(0,I)}
    \left[
    \left\|
    v_t(\psi_t(x_0);\theta)
    -
    \left(x_1-(1-\sigma_{\min})x_0\right)
    \right\|^2
    \right].
    \label{eq:ot_cfm_loss}
\end{equation}
Thus, flow matching with OT conditional vector fields trains a neural ODE to match a simple straight-line conditional transport from Gaussian noise to data samples. After training, generation is performed by sampling $x_0\sim\mathcal{N}(0,I)$ and numerically solving Equation \ref{eq:cnf_ode} from $t=0$ to $t=1$ using the learned vector field $v_t(x;\theta)$.

More generally, flow matching is not restricted to a Gaussian prior distribution \cite{tong2024improving}. Let $\rho_0(x_0)$ denote an arbitrary source distribution and let $q(x_1)$ denote the target data distribution. Instead of independently sampling $x_0$ and $x_1$, one may introduce a coupling
\begin{equation}
\pi(x_0,x_1)\in \Pi(\rho_0,q),
\label{eq:general_coupling}
\end{equation}
where $\Pi(\rho_0,q)$ denotes the set of joint distributions whose marginals are $\rho_0$ and $q$, respectively. Given a pair $(x_0,x_1)\sim \pi$, a simple deterministic conditional path can be defined by the linear interpolation
\begin{equation}
x_t=\psi_t(x_0,x_1)=(1-t)x_0+t x_1,
\qquad t\in[0,1],
\label{eq:general_linear_path}
\end{equation}
which induces the conditional distribution
\begin{equation}
p_t(x\mid x_0,x_1)
=
\delta\left(x-\psi_t(x_0,x_1)\right).
\label{eq:general_conditional_path}
\end{equation}
The corresponding conditional vector field is simply the time derivative of the path,
\begin{equation}
u_t(x_t\mid x_0,x_1)
=
\frac{d}{dt}\psi_t(x_0,x_1)
=
x_1-x_0.
\label{eq:general_conditional_vf}
\end{equation}
Therefore, the generalized conditional flow matching objective can be written as
\begin{equation}
\mathcal{L}_{\mathrm{GCFM}}(\theta)
=
\mathbb{E}_{t\sim\mathcal{U}[0,1],(x_0,x_1)\sim\pi}
\left[
\left\|
v_t\left((1-t)x_0+t x_1;\theta\right)
-
(x_1-x_0)
\right\|^2
\right].
\label{eq:gcfm_loss}
\end{equation}
This formulation shows that the source distribution $\rho_0$ does not need to be Gaussian. The only requirement is that samples can be drawn from $\rho_0$ and that a suitable coupling $\pi$ between $\rho_0$ and $q$ can be constructed. In practice, $\pi$ may be chosen as the independent coupling $\rho_0(x_0)q(x_1)$, which pairs source and target samples in a way that reduces the intersection probability and transport cost and yields straighter probability paths.

After training, sampling is performed by first drawing a source sample $x_0\sim\rho_0$ and then integrating the learned neural ODE from $t=0$ to $t=1$:
\begin{equation}
\hat{x}_1
=
\phi_1(x_0)
=
x_0+\int_0^1 v_t(\phi_t(x_0);\theta),dt,
\qquad
\frac{d}{dt}\phi_t(x_0)=v_t(\phi_t(x_0);\theta),
\quad
\phi_0(x_0)=x_0.
\label{eq:general_sampling}
\end{equation}
For example, using an Euler solver with $N$ time steps, the sampling process can be written as
\begin{equation}
x_{k+1}=x_k+\Delta t\,v_{t_k}(x_k;\theta),
\qquad
t_k=\frac{k}{N},\quad
\Delta t=\frac{1}{N},\quad
k=0,\ldots,N-1,
\label{eq:euler_sampling}
\end{equation}
where $x_0\sim\rho_0$ and $x_N$ is the generated sample. 

As can be observed from Equation \ref{eq:general_sampling} and \ref{eq:euler_sampling}, one-step or few-step generation becomes feasible when the learned velocity field induces nearly straight transport trajectories. In standard flow matching, however, multiple conditional training paths may pass through the same or nearby regions of the sample space with different target velocities. As a result, the learned marginal velocity field reflects an averaged transport direction, which can cause the generated trajectories to deviate from ideal straight transports. Consequently, multi-step sampling is often needed to better integrate the learned flow toward the target distribution.

\subsubsection{One-Step Generation}\

Since achieving one-step generation is crucial for improving model efficiency, several related studies have been proposed; however, these methods still suffer from significant limitations or costs. Rectified Flow \cite{liu2023flow} performs iterative rectification to reduce trajectory curvature, thereby enabling efficient few-step generation. However, because reflow relies on samples generated by the model itself from the previous round, it may solidify or even amplify the biases introduced by the preceding model. More recently, MeanFlow \cite{geng2026mean} introduced a framework that learns mean velocities rather than instantaneous velocities, improving the quality of few-step or even one-step flow-based generation. Nevertheless, this method transfers much of the computational cost to the training stage, resulting in slower training and more difficult hyperparameter tuning. In addition, most existing methods can substantially reduce the number of sampling steps, but achieving high-quality one-step generation remains highly challenging.

It is worth noting that existing flow matching models commonly adopt a Gaussian distribution as the source distribution; however, flow matching itself does not inherently rely on Gaussian sources \cite{tong2024improving}. Unlike natural video synthesis, scientific simulation tasks often contain useful prior information. For example, von Mises stress fields are non-negative, and the stress values must be zero in regions without material. This allows us to construct a prior-informed random distribution as the source distribution for flow matching, which not only preserves stochasticity to the model but also substantially reduces the likelihood of crossing conditional transport paths during training, thereby enabling one-step generation.

Based on the simulation problem considered in this work, we construct such an empirical distribution $p_{\mathrm{geo}}$, which is induced by the stochastically generated geometry samples. Specifically, we use pixels to represent plates with random holes; the pixel values corresponding to the holes are set to 0, while those corresponding to the material region are set to half of the maximum value. In the implementation, since the stress fields are normalized to the range $[-1,1]$, these two values are set to -1 and 0, respectively. Therefore, each binary geometry field can be represented as a binary-valued tensor $\mathbf{x}_0\in\{-1,0\}^{T\times H\times W}$. Here, $T$, $H$, and $W$ denote the number of frames, height, and width of each sample, respectively. In this research, the number of frames is equal to the number of loading-unloading steps. Let $\mathcal{D}_{\mathrm{geo}}=\{\mathbf{x}_0^{(k)}\}_{k=1}^{N}$ denote the set of geometry fields generated in this manner. The corresponding geometry source distribution is defined as the empirical distribution
\begin{equation}
    p_{\mathrm{geo}}(\mathbf{x}_0)
    =
    \frac{1}{N}
    \sum_{k=1}^{N}
    \delta\!\left(\mathbf{x}_0-\mathbf{x}_0^{(k)}\right),
    \qquad
    \mathbf{x}_0^{(k)}\in\{-1,0\}^{T\times H\times W},
    \label{eq:p_geo}
\end{equation}
where $\delta(\cdot)$ denotes the Dirac delta distribution. During training, each geometric source sample is paired one-to-one with its corresponding target stress field sample. This coupling can reduce the likelihood of conditional path intersection. The reason is that, when the hole locations are properly matched, the dimensions corresponding to pixels inside the holes contribute zero distance between the paired source and target samples. In contrast, if the source and target samples are mismatched, the hole regions may be spatially misaligned, causing zero-valued pixels in the holes of one sample to be paired with nonzero pixels in the other. This misalignment increases the contribution to the L2 distance and therefore makes the overall distance more likely to be larger than that of a properly matched pair. To demonstrate why this property can reduce the probability of intersections, we consider the following Lemma 1.

\par\indent\textbf{Lemma 1.}
Consider four distinct points $x_i,x_j,y_i,y_j\in\mathbb{R}^n$ equipped with the Euclidean metric and $i \neq j$. If
\begin{equation}
d(x_i,y_i) < d(x_i,y_j),
\end{equation}
and
\begin{equation}
d(x_j,y_j) < d(x_j,y_i),
\end{equation}
then the open line segments $\overline{x_i y_i}$ and $\overline{x_j y_j}$ do not
intersect.

Lemma 1 can be readily proved using the triangle inequality, and the proof is provided in \ref{app:proof}. Therefore, we can conduct a series of experiments to compare the likelihood of intersection between line segments connecting two randomly selected source-target pairs, i.e., the conditional paths constructed during training, when using the geometry-based empirical distribution and the Gaussian, respectively. We randomly select two samples from the dataset and compute the distances between these two samples and their corresponding geometric source points. We then determine whether the line segments connecting the two pairs intersect by checking whether the condition in Lemma 1 is satisfied. This experiment is repeated $10{,}000$ times. We perform the same experiment using a Gaussian distribution as the source distribution, also with $10{,}000$ random trials, and compare the statistical results of the two cases. It should be noted that, since this work employs a VAE for dimensionality reduction, the data samples must first be encoded into the latent space by the VAE before computing the distances. In addition, because the samples considered in this work are extremely high-dimensional, the distance computation is computationally expensive; therefore, only $10{,}000$ trials are conducted for each case.

Our experimental results show that, when starting from the geometry-based empirical distribution, $99.96\%$ of the trials satisfy Lemma 1. In contrast, when starting from the Gaussian distribution, only $5.65\%$ of the trials satisfy this condition. It should be emphasized that failure to satisfy Lemma 1 only indicates that the trajectories may intersect, rather than that they necessarily intersect. Conversely, satisfying Lemma 1 mathematically guarantees that the corresponding trajectories cannot intersect. These results demonstrate that, when training is performed from the geometry-based empirical distribution, the probability of conditional path intersections in the sample space is extremely low. This encourages shorter and less conflicting conditional transport paths, resulting in a learned vector field whose sampling trajectories are closer to straight lines. As a result, the proposed model can perform one-step generation with satisfactory performance. Furthermore, this approach directly incorporates geometric information into the model, thereby avoiding the need for a dedicated CNN-based feature extractor and the subsequent geometry conditioning injection procedure commonly employed in previous studies \cite{jadhav2023stressd, kota2026hybrid}.

\subsection{DiT Backbone}\

\begin{figure}[htp]
    \centering
    \includegraphics[width=\linewidth]{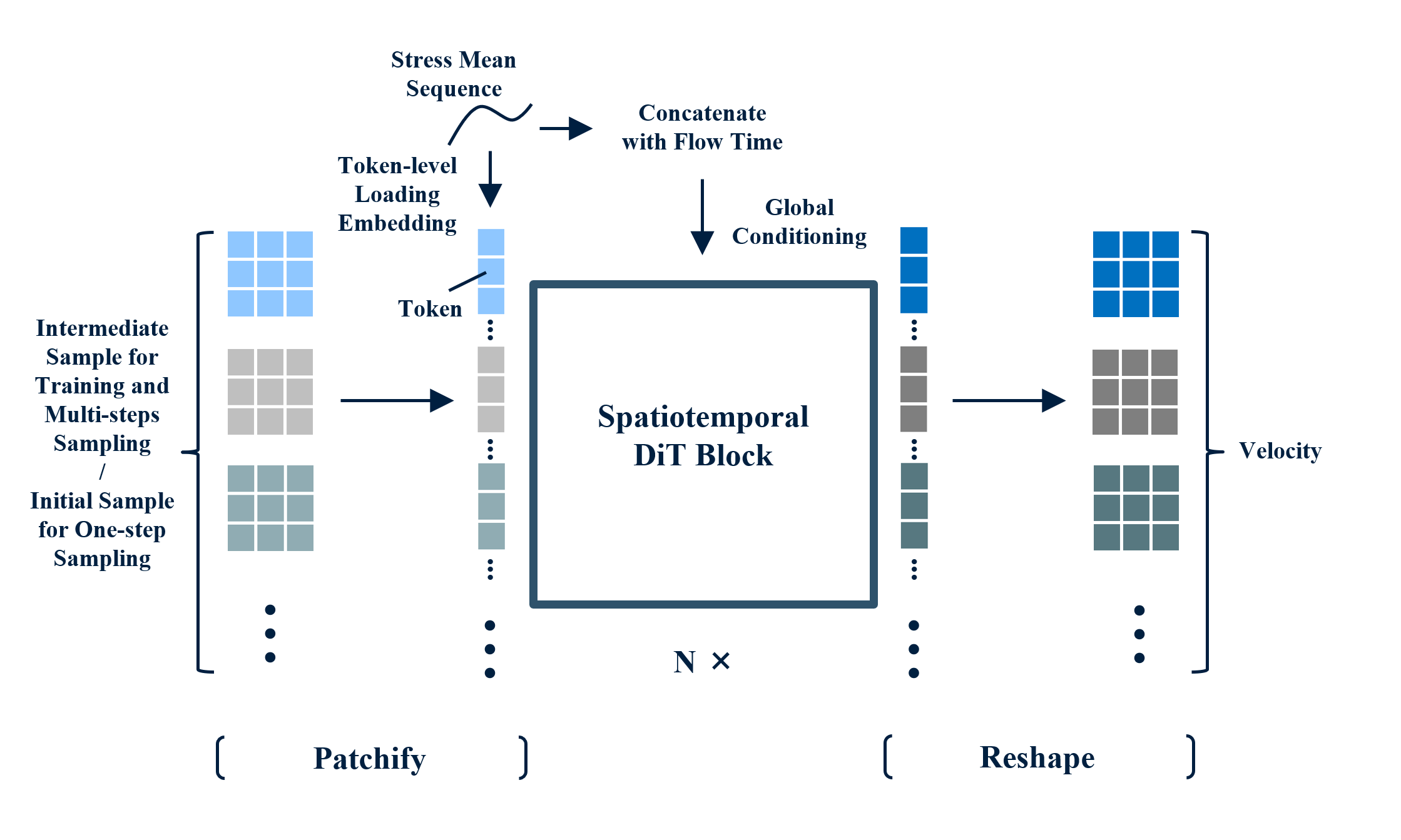}
    \caption{Spatiotemporal DiT backbone. After being compressed into the latent space, each frame is patchified, and each patch is projected into a token. The frame-wise spatial tokens are then concatenated across time to form a spatiotemporal token sequence. Positional encoding and weighted token-level loading embedding are added to the tokens. The resulting sequence is processed by DiT blocks, where timestep and loading-path conditions modulate the blocks through adaLN. Finally, the processed tokens are projected back to latent patches and unpatchified to recover the original spatiotemporal latent shape.}
    \label{fig:Backbone}
\end{figure}

The main network in this work, DiT \cite{peebles2023scalable}, was originally introduced as a backbone for diffusion models in the image synthesis task, from which its name is derived. Recently, this type of backbone has been widely adopted in flow matching and is gradually becoming a mainstream approach for image and video generation. We present the details of the processing pipeline of the backbone used in this research in Figure \ref{fig:Backbone}, including patchification and reshaping, the token-level loading embedding, and the global conditioning. The architecture of DiT is similar to that of ViT \cite{dosovitskiy2020image}. For image synthesis tasks, DiT first patchifies the image and flattens patches into vectors, which are then projected as tokens and processed by the transformer to learn the spatial dependency. In this work, the adopted spatiotemporal DiT patchifies each frame, and each patch is projected into a token. The frame-wise spatial tokens are then concatenated across time to form a spatiotemporal token sequence, depending on fixed spatial and temporal order. Positional encoding and weighted token-level loading embedding are added to the tokens. The resulting sequence is processed by DiT blocks. This design enables the model to directly capture spatiotemporal consistency while jointly accounting for the effects of geometry and the loading path, which is crucial for the simulation of plastic stress fields.

It is worth noting that, in this work, the DiT backbone operates in the latent space of a Scientific VAE. Specifically, the VAE compresses the original fields from pixel space into the latent space frame by frame for processing by the DiT, and then reconstructs the latent representations back into the pixel space frame by frame after the DiT processing is completed. This design is motivated by the quadratic computational complexity of the self-attention mechanism with respect to the sequence length, i.e., \(\mathcal{O}(N^2)\). For high-resolution images or videos, the resulting number of tokens can become prohibitively large, leading to excessive computational costs and GPU VRAM limitations. Therefore, dimensionality reduction through a VAE is essential. To construct the embedded loading information, we employ a ViT-Transformer network to predict the mean stress sequence, which helps alleviate the burden on the main network in learning the constitutive response. If the ViT-Transformer is not used, the loading information is instead obtained from the displacement sequence. Moreover, both the VAE and the ViT-Transformer network are pretrained; that is, during the training of the main network, the parameters of these two auxiliary networks are fixed. The details about auxiliary networks are discussed in the Subsection \ref{subsec:auxiliary_networks}.

\subsubsection{Token-Level Loading Embedding}\

The model in this research is designed to handle arbitrary loading sequences. Our study shows that, due to the high uncertainty of the loading sequence and the complexity of the constitutive model, directly injecting the loading information through global conditioning is difficult for the model to effectively learn the loading state. As a result, to enhance the model's ability to learn the evolution of stress fields under varying loading conditions, we propose a token-level loading information embedding that works in conjunction with global conditioning, which is shown in Equation \ref{eq:tllembedding}. Here, $\mathbf{x}_{\mathrm{in}}$ and $\mathbf{x}$ denote the token before and after encoding, respectively. $\mathbf{P}$ is the positional encoding associated with the current token, $\mathbf{L}$ represents the loading information of the frame to which the current token belongs, and $\eta$ is a learnable parameter.

\begin{equation}\label{eq:tllembedding}
\mathbf{x} = \mathbf{x}_{\mathrm{in}} + \mathbf{P} + \eta \mathbf{L}
\end{equation}

Specifically, we first project the loading information of the current frame into the embedding dimension. After being multiplied by a learnable parameter $\eta$, the loading embedding is broadcast to every token, i.e., every patch, within that frame, and is then added to the token together with the positional encoding. The learnable parameter $\eta$ controls the strength of the loading embedding, thereby preventing it from interfering with the positional encoding. It should be noted that the entire loading path is also concatenated with the flow time and injected through global conditioning to help the model capture global dependencies. Here, the flow time belongs to $[0,1]$, which represents the time associated with the motion of data points in the sample space, rather than the loading time step.

\subsubsection{DiT Block}\

The DiT block adopted in this work follows the standard adaLN-Zero design, in which the conditioning information is injected into each transformer block through adaptive layer normalization and residual gating. Given the input token sequence $\mathbf{x}\in\mathbb{R}^{N\times d}$ and a global conditioning vector $\mathbf{c}$, the conditioning vector is first processed by an MLP to generate a set of modulation parameters, including the shift and scale parameters $(\boldsymbol{\beta}_1,\boldsymbol{\gamma}_1)$ and $(\boldsymbol{\beta}_2,\boldsymbol{\gamma}_2)$ for the attention and feed-forward sublayers, respectively, as well as the corresponding residual scaling factors $\alpha_1$ and $\alpha_2$. These parameters are used to adaptively modulate the normalized token features before they are passed into the multi-head self-attention and pointwise feed-forward modules.

Specifically, the first sublayer applies layer normalization to the input tokens and then performs a conditioning-dependent affine transformation:
\begin{equation}
\label{eq:adaln-attn-modulation}
\tilde{\mathbf{x}}_1
=
(1+\boldsymbol{\gamma}_1)\odot \mathrm{LN}(\mathbf{x})
+
\boldsymbol{\beta}_1
\end{equation}
where $\odot$ denotes element-wise multiplication and LN denotes layer normalization. The modulated tokens are then processed by the multi-head self-attention module, and the resulting features are added back to the input through a gated residual connection:
\begin{equation}
\label{eq:adaln-attn-residual}
\mathbf{x}'
=
\mathbf{x}
+
\alpha_1
\mathrm{MSA}(\tilde{\mathbf{x}}_1)
\end{equation}
here MSA denotes a multi-head self-attention layer. The second sublayer follows the same principle. The intermediate tokens $\mathbf{x}'$ are normalized and modulated using another set of conditioning-dependent parameters:
\begin{equation}
\label{eq:adaln-ffn-modulation}
\tilde{\mathbf{x}}_2
=
(1+\boldsymbol{\gamma}_2)\odot \mathrm{LN}(\mathbf{x}')
+
\boldsymbol{\beta}_2
\end{equation}
The modulated features are then passed through a pointwise feed-forward network and added to the residual stream with another learnable gate:
\begin{equation}
\label{eq:adaln-ffn-residual}
\mathbf{x}_{\mathrm{out}}
=
\mathbf{x}'
+
\alpha_2
\mathrm{FFN}(\tilde{\mathbf{x}}_2)
\end{equation}
where FFN denotes a feed-forward network.

This design allows the conditioning information to control both the feature normalization and the strength of the residual updates in each DiT block. The shift and scale parameters determine how the token features are modulated before attention and feed-forward processing, while the residual scaling factors $\alpha_1$ and $\alpha_2$ regulate the contribution of each sublayer to the output tokens. In the adaLN-Zero formulation, these residual pathways are initialized to be close to zero, making the block behave approximately as an identity mapping at the beginning of training. This improves optimization stability and enables the model to gradually learn how the conditioning information should influence the token representations. The global conditioning vector $\mathbf{c}$ is concatenated by the flow time and the loading information sequence in this work.

\subsection{Auxiliary Networks}\label{subsec:auxiliary_networks}

\subsubsection{Scientific VAE}\

Since the model developed in this work is designed to generate high-resolution stress fields and involves the simultaneous generation of multiple frames, i.e., a video synthesis task, a VAE is required for dimensionality reduction. The variational autoencoder employed in this study is obtained by modifying and fine-tuning an off-the-shelf VAE \cite{Rombach_2022_CVPR} that is widely used in the AI community. The original VAE takes RGB images as input and has an encoder with a spatial downsampling factor of 8. Specifically, it maps an input tensor of size $\left(3,256,256\right)$ to a latent tensor of size $\left(4,32,32\right)$. However, this VAE was originally trained on natural images, and its loss function was designed to emphasize perceptual similarity from the perspective of human vision. Therefore, it cannot be directly applied to stress-field simulation, and both architectural modifications and fine-tuning are required.

First, since each frame of our stress-field data contains only one channel, we add one adapter to each end of the original VAE to match the number of input and output channels. We then construct a loss function tailored for scientific fields, as defined in Equation \ref{eq:vae-loss}, to fine-tune the model into a scientific VAE. 
\begin{equation}
\label{eq:vae-loss}
\mathcal{L}_{\mathrm{total}}
=
\mathcal{L}_{\mathrm{MSE}}
+
w_{1}\mathcal{L}_{L1}
+
w_{2}\mathcal{L}_{\mathrm{grad}}
+
w_{3}\mathcal{L}_{\mathrm{KL}} .
\end{equation}
Here, $\mathcal{L}_{\mathrm{grad}}$ and $\mathcal{L}_{\mathrm{KL}}$ denote the gradient loss and the Kullback-Leibler divergence loss, respectively. The term $\mathcal{L}_{\mathrm{MSE}}$ serves as the primary reconstruction term and strongly penalizes large pixel-wise errors. The auxiliary reconstruction term $\mathcal{L}_{L1}$ is introduced to further reduce the overall reconstruction bias. The gradient loss $\mathcal{L}_{\mathrm{grad}}$ helps the model capture variations in high-gradient regions, such as areas near geometric changes and regions with sharp stress variations. The KL divergence term $\mathcal{L}_{\mathrm{KL}}$ regularizes the latent space and encourages it to be a more continuous and structured latent representation. The weight coefficients $w_{1}$, $w_{2}$, and $w_{3}$ are used to balance the contributions of the corresponding loss terms. Using the above loss function, we fine-tune the scientific VAE on our stress-field dataset.

\subsubsection{ViT-Transformer}\

The ViT-Transformer \cite{zhou2025vit} was originally developed for constitutive modeling of composites with random fiber distributions, with the aim of accelerating simulations. This network consists of a ViT encoder that processes image data to extract geometric features. The extracted geometric information is then concatenated with the microscopic strain sequence and passed to a transformer decoder to predict the microscopic stress response. In this work, we introduce this network to predict the mean stress sequence from the geometric image and the displacement sequence. The motivation is that, according to our study, under limited model and dataset sizes, directly injecting the displacement sequence causes the model to miss the overall stress intensity in minority frames. This is mainly due to the complexity of the plastic constitutive behavior and the fact that our loading-unloading paths are fully random. The purpose of introducing the ViT-Transformer is therefore to assist the main network in learning how the constitutive response affects the evolution of the global stress magnitude.

It should be noted that, motivated by the industrial demand for reducing computational cost and the practical scarcity of large-scale datasets, this work focuses on achieving the best possible model performance under reasonable model and dataset sizes. If the number of trainable parameters and the dataset size were substantially increased, and if extensive GPU resources were available for training, the auxiliary network would, in principle, not be necessary, since the original loading data already contains all the required information.

\subsection{Implementation Details}\

First, we pretrain the auxiliary networks, i.e., the scientific VAE and the ViT-Transformer. After convergence, the parameters of the auxiliary networks are frozen, and the training of the main DiT network is then performed. The hyperparameters of the DiT network are summarized in Table \ref{tab:hp_DiT} and the number of trainable parameters is 133 million. For the training of both the main and auxiliary networks, we adopt stochastic gradient-based optimization with the Adam optimizer. Min-max normalization is applied throughout the training process. When training the main DiT network and the ViT-Transformer, we use the MSE loss; however, their learning targets are different. The DiT is trained to predict the velocity field in sample space, whereas the ViT-Transformer directly fits the mean stress sequence. In contrast, the scientific VAE is trained using the hybrid loss defined in Equation \ref{eq:vae-loss}.

\begin{table}[h]
\centering
\caption{Hyperparameters of spatiotemporal DiT network.}
\label{tab:hp_DiT}
\renewcommand{\arraystretch}{1.2}
\setlength{\tabcolsep}{16pt}
\begin{tabular}{l l l}
\hline
Hyper-parameters \\ \hline
Depth & 12 \\ 
Hidden Size & 768 \\ 
Patch Size & 2 \\ 
Number of Heads & 12 \\ 
MLP Ratio & 4 \\ \hline
\end{tabular}
\end{table}

In this work, the training of the main DiT network and the pretraining of the scientific VAE are both performed on a single NVIDIA H100 SXM (80 GB) GPU, while the pretraining of the ViT-Transformer and model sampling on the GPU are conducted using a single NVIDIA GeForce RTX 5080 (16 GB) GPU. For the efficiency comparison between the proposed model and FEM, all evaluations are performed using a single Intel(R) Core(TM) Ultra 7 265K CPU.

\section{Datasets}\label{sec:Datasets}

Our dataset consists of two-dimensional von Mises stress fields of square plates generated from stochastic loading-unloading paths applied to random geometries, represented in pixel form. The material behavior is elastoplasticity governed by a J2 plasticity constitutive model. The simulations are performed using ABAQUS, and the finite element results are converted into pixel-based representations using inverse distance weighting interpolation. The dataset contains a total of 20,000 samples, with each sample consisting of 20 frames. The resolution of each frame is $256 \times 256$. During dataset generation, the samples are divided into three groups. The first group contains 10,000 samples, where each sample includes a single circular hole with a varied radius and a random position. The second group contains 5,000 samples, where each sample contains 3 holes with random positions. The third group contains 5,000 samples, where each sample contains 6 holes with random positions. For each group, 500 samples are reserved for testing.

\begin{figure}[htp]
    \centering
    \includegraphics[width=6cm]{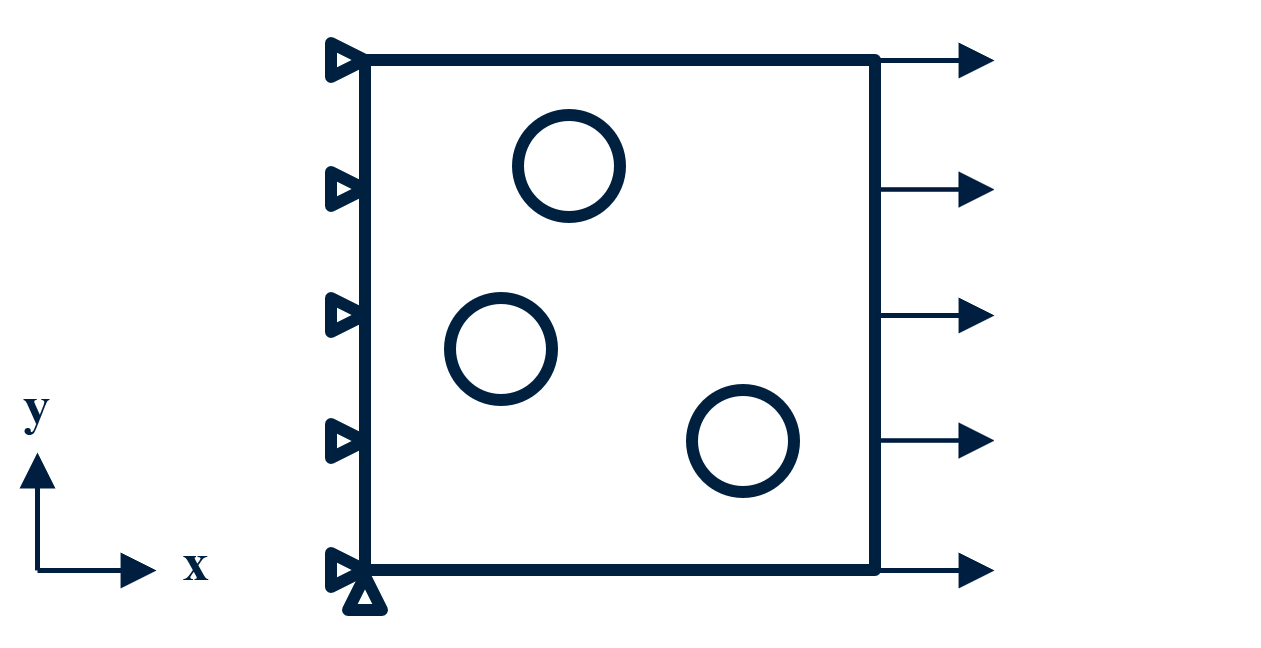}
    \caption{Boundary conditions and loading. The right side boundary is displaced in the direction that depends on the sign of the current step of stochastic loading-unloading paths. If the sign of the displacement is negative, the arrow points to the left.}
    \label{fig:BCs}
\end{figure}

The detailed generation procedure is as follows. First, we generate a random displacement sequence containing all 20,000 samples. The sequence-generation strategy is modified from the random-walk-based method proposed in \cite{zhou2025machine}, so as to ensure that a large number of integration points enter the plastic range. We then generate three groups of geometric datasets with random holes and merge them into a unified dataset. The geometric and loading data are subsequently imported into ABAQUS CAE through preprocessing scripts, from which the corresponding input files are generated. The side length of the square plate is set to a unit length, and the boundary conditions and loading are illustrated in Figure \ref{fig:BCs}. It should be noted that the direction of the arrows on the right boundary depends on the sign of the displacement at the current loading or unloading step. The arrows point to the right, as shown in the figure, only when the displacement is positive. The displacement of the left boundary is fixed in the $x$-direction while remaining free to slide in the $y$-direction. A displacement load is applied to the right boundary, and the displacement magnitude at each time step is determined by the pre-generated random displacement path. In addition, the displacement of the lower-left corner point is constrained in the $y$-direction to prevent rigid-body motion. It should be noted that this is a relatively small dataset for generative modeling. In comparison, the diffusion-model-based work in \cite{jadhav2023stressd} used a dataset containing more than 120,000 samples, even though it only modeled linear elastic stress fields for 2D cantilever structures.

\section{Results}\label{sec:Results}

In this section, we systematically evaluate the performance and efficiency of the proposed model and visualize the generated samples. Meanwhile, the evaluation includes a comparison between one-step and multi-step generation in terms of both accuracy and computational efficiency, as well as a comparison between the proposed model and the finite element method. To compare the difference between the sample $\hat{\mathbf{x}_1}$ generated by our model and the ground-truth target point $\mathbf{x}_1$,  the evaluation metrics are defined as follows:
\begin{equation}
\mathrm{MAE}
=
\frac{1}{N}
\left\|
\mathbf{x}_1 - \hat{\mathbf{x}}_1
\right\|_1 ,
\label{eq:mae}
\end{equation}

\begin{equation}
\mathrm{RMAE}
=
\frac{\mathrm{MAE}}
{\max(\mathbf{x}_1) - \min(\mathbf{x}_1)}
\times 100\% .
\label{eq:rmae}
\end{equation}
Here, MAE denotes the mean absolute error, which measures the overall error level of the generated samples. RMAE denotes the relative mean absolute error, which quantifies the absolute error of the model relative to the overall magnitude of the ground-truth field. When plotting the error maps using relative errors, the MAE in the denominator of Equation \ref{eq:rmae} only needs to be replaced by the value of each pixel.

\subsection{Performance}\

In this section, we evaluate the performance and generalization capability of the proposed model using our test dataset, in which both the geometries and loading paths are unseen during training. The geometries and loading paths used in the test set are randomly generated. All samples presented in this section are generated by the model using only one sampling step. Due to space limitations in the main text, it is infeasible to display all frames of each sample. Therefore, we consistently select the 15th frame for visualization in this section. In the \ref{app:samples}, we provide frame by frame visualizations of the complete samples, together with comparisons against the corresponding ground truth. We also present samples generated using 20 sampling steps in \ref{app:samples} to facilitate comparison with the one-step generation results.

\begin{figure}[htbp]
    \centering
    \begin{subfigure}{\textwidth}
        \centering
        \makebox[\textwidth][l]{%
            \textbf{(a)}\quad
            \includegraphics[width=0.96\textwidth]{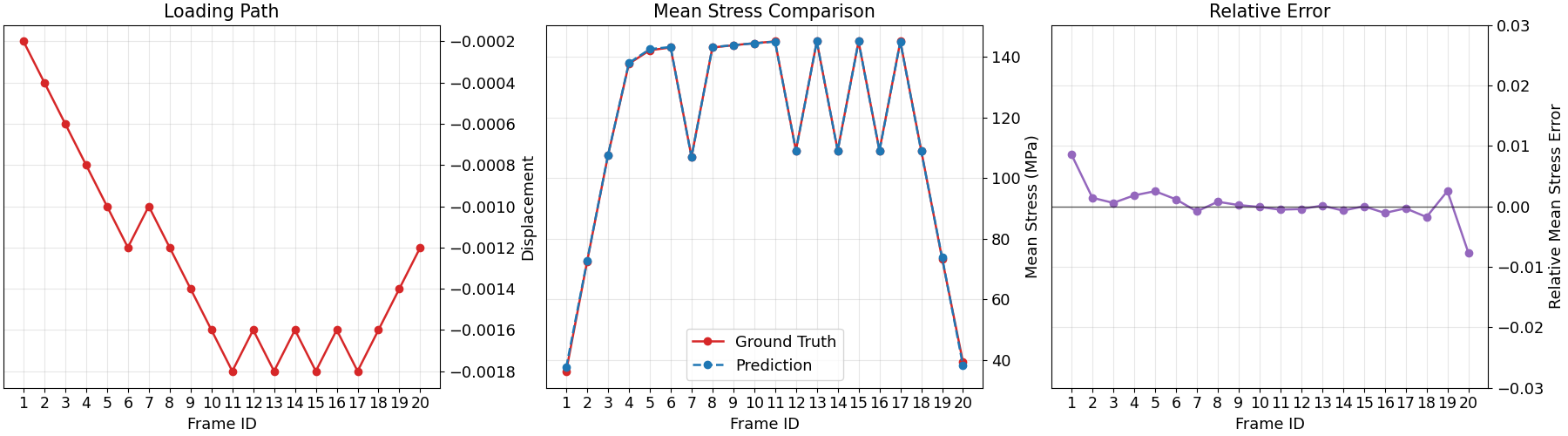}
        }
        \phantomsubcaption
        \label{fig:MeanStress_1hole_1step}
    \end{subfigure}
    \vspace{0.5em}
    \begin{subfigure}{\textwidth}
        \centering
        \makebox[\textwidth][l]{%
            \textbf{(b)}\quad
            \includegraphics[width=0.96\textwidth]{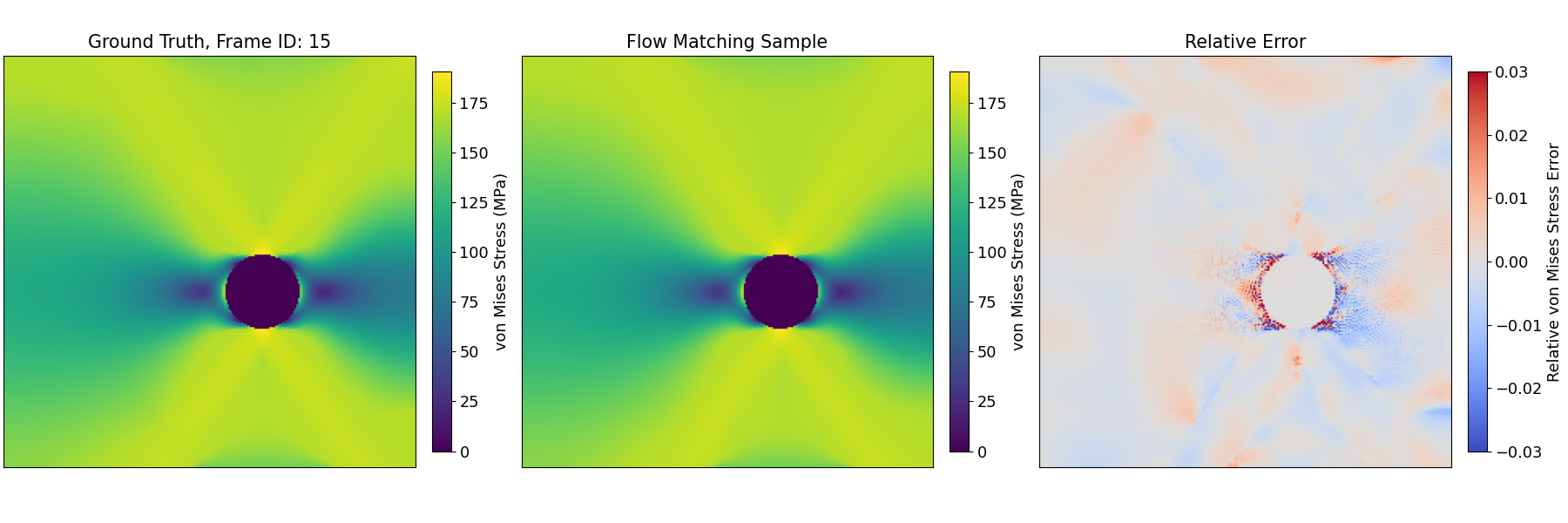}
        }
        \phantomsubcaption
        \label{fig:Sample_1hole_1step_frame}
    \end{subfigure}
    \caption{Performance of one-step generation for the single hole case.}
    \label{fig:1hole_1step}
\end{figure}

\begin{figure}[htbp]
    \centering
    \begin{subfigure}{\textwidth}
        \centering
        \makebox[\textwidth][l]{%
            \textbf{(a)}\quad
            \includegraphics[width=0.96\textwidth]{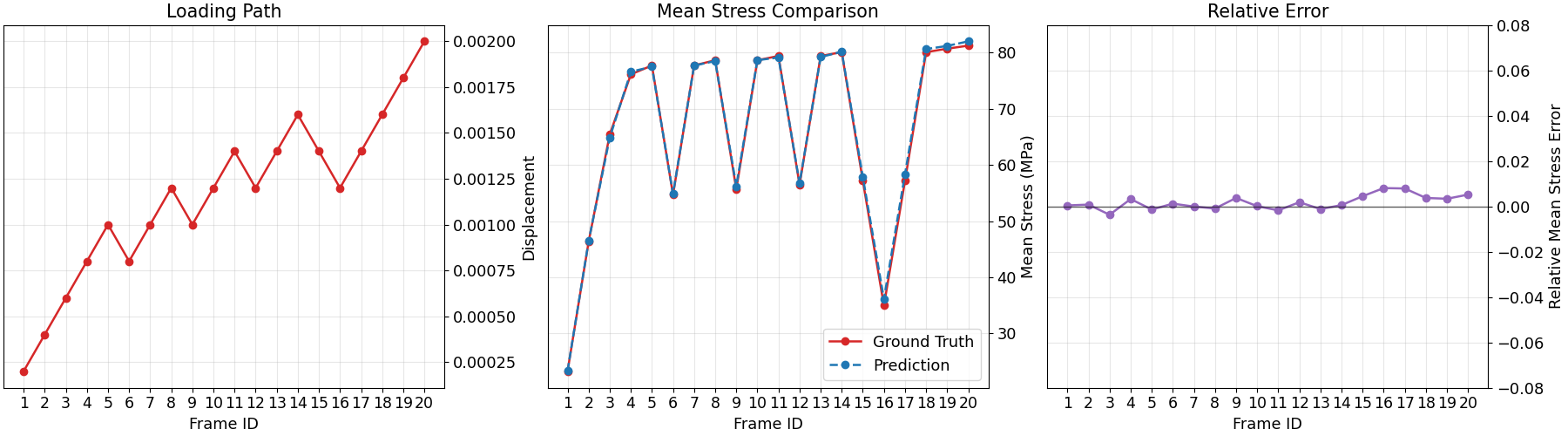}
        }
        \phantomsubcaption
        \label{fig:MeanStress_3holes_1step}
    \end{subfigure}
    \vspace{0.5em}
    \begin{subfigure}{\textwidth}
        \centering
        \makebox[\textwidth][l]{%
            \textbf{(b)}\quad
            \includegraphics[width=0.96\textwidth]{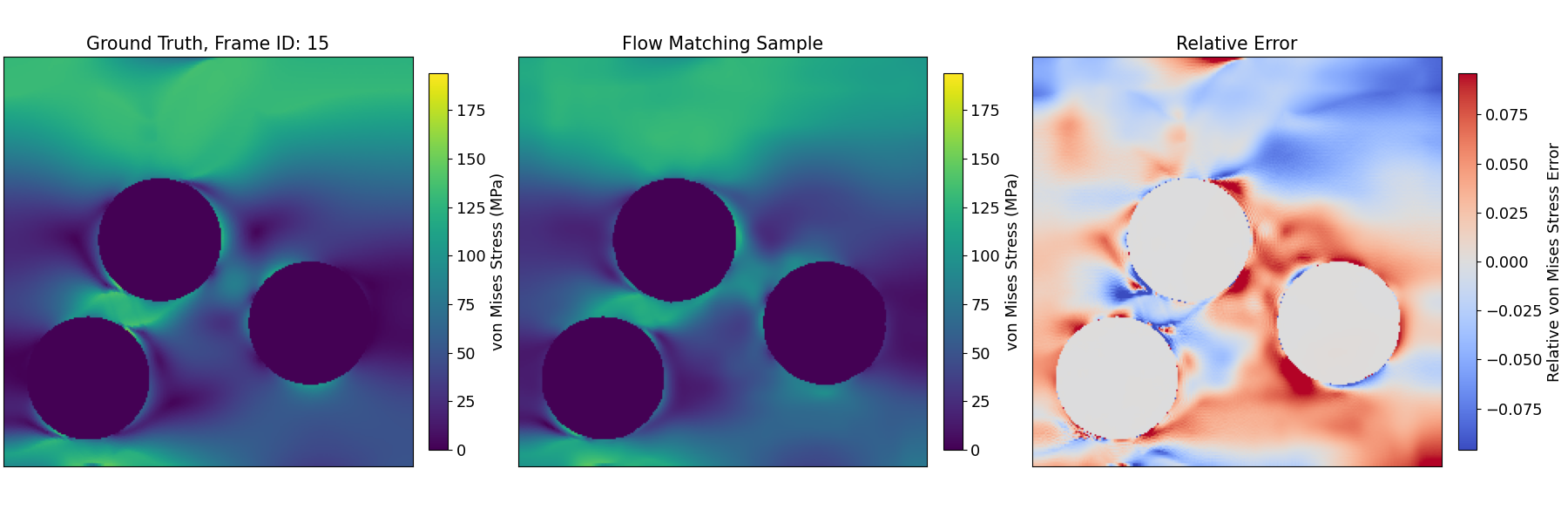}
        }
        \phantomsubcaption
        \label{fig:Sample_3holes_1step_frame}
    \end{subfigure}
    \caption{Performance of one-step generation for the 3 holes case.}
    \label{fig:3holes_1step}
\end{figure}

\begin{figure}[htbp]
    \centering
    \begin{subfigure}{\textwidth}
        \centering
        \makebox[\textwidth][l]{%
            \textbf{(a)}\quad
            \includegraphics[width=0.96\textwidth]{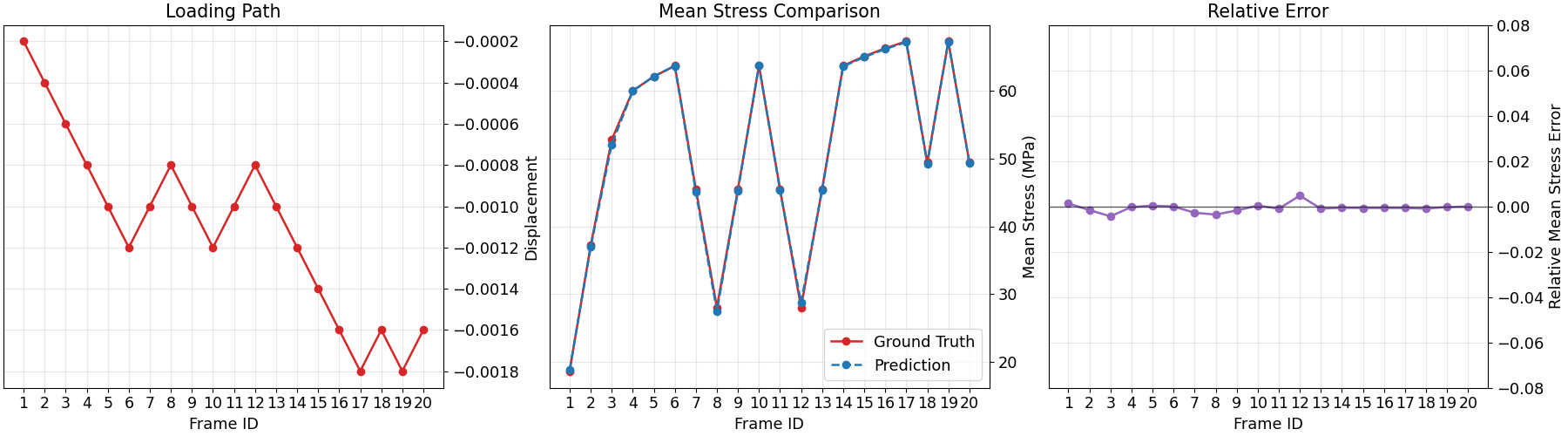}
        }
        \phantomsubcaption
        \label{fig:MeanStress_6holes_1step}
    \end{subfigure}
    \vspace{0.5em}
    \begin{subfigure}{\textwidth}
        \centering
        \makebox[\textwidth][l]{%
            \textbf{(b)}\quad
            \includegraphics[width=0.96\textwidth]{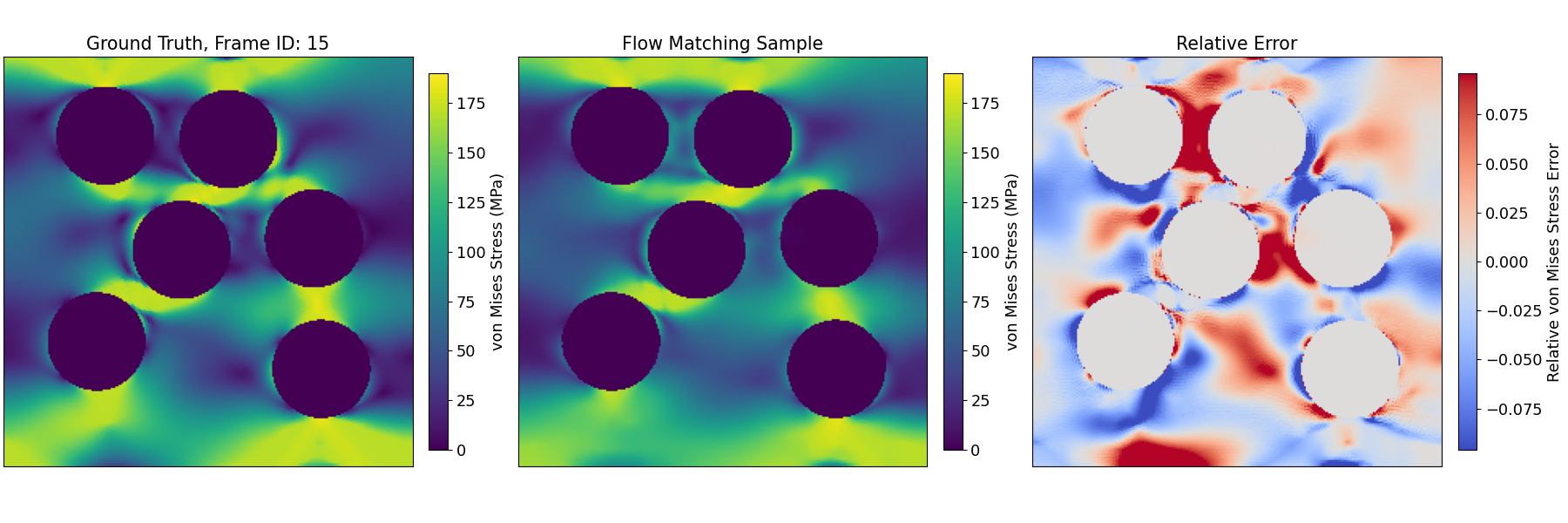}
        }
        \phantomsubcaption
        \label{fig:Sample_6holes_1step_frame}
    \end{subfigure}
    \caption{Performance of one-step generation for the 6 holes case.}
    \label{fig:6holes_1step}
\end{figure}

Figure \ref{fig:1hole_1step} shows the one-step generation results for the single hole case. Figure \ref{fig:MeanStress_1hole_1step} presents the corresponding loading path. In addition, we compute the spatial average of the model prediction at each frame and compare it with the ground-truth average. This plot clearly demonstrates that the model accurately learns the influence of loading path and plastic behavior on the global magnitude of the stress field. Figure \ref{fig:Sample_1hole_1step_frame} shows that the proposed model can accurately capture both the spatial distribution and the intensity of the stress field, even after the field has undergone many random loading and unloading steps. Not only is the hole location accurately captured, but the stress distribution and magnitude are also predicted with high accuracy both inside the hole, outside the hole, and around the hole boundaries. Figures \ref{fig:3holes_1step} and \ref{fig:6holes_1step} show the one-step generation results for the 3 holes case and 6 holes case, respectively. It can be observed that both the overall evolution of the stress magnitude and the spatial distribution of the stress field are accurately captured by the proposed model. Although the accuracy is slightly lower than that of the single hole case, this is reasonable because the data points with 3 holes and 6 holes are each only half of those for the single hole points in the dataset, and the stress distributions become more complex in the presence of multiple holes. Furthermore, from the perspective of the error distribution, the model exhibits excellent performance in high-value regions. This is particularly advantageous for simulation tasks, as the behavior in high-stress regions is usually of primary interest. The dominant errors are mainly concentrated in regions with high stress gradients.

Table \ref{tab:performance} reports the average MSE and RMSE of the model under different numbers of sampling steps and across different types of test datasets. The results show that our model maintains high accuracy even when evaluated on unseen geometries and loading-unloading paths, demonstrating strong generalization capability. In addition, the results indicate that the error of one-step generation is comparable to that of multi-step generation, suggesting that satisfactory results can be obtained using only a single sampling step. Meanwhile, when maximum inference speed is not the primary objective, the model accuracy can be further improved by slightly increasing the number of sampling steps. Specifically, increasing the sampling steps to 5 reduces the MSE by 12\% and 16\% for the single-hole and multi-hole cases, respectively.

\begin{table}[h]
\centering
\caption{Performance evaluation of proposed framework.}
\label{tab:performance}
\renewcommand{\arraystretch}{1.2}
\setlength{\tabcolsep}{16pt}
\begin{tabular}{c c c c}
\hline
Number of sampling steps & Metrics & Single hole & Multi holes \\ \hline
\multirow{2}{*}{One-step} 
& MAE & 1.05 & 6.41 \\ 
& RMAE & 0.476 \% & 2.90 \% \\ [6pt]
\multirow{2}{*}{5 steps} 
& MAE & 0.924 & 6.16 \\ 
& RMAE & 0.419 \% & 2.79 \% \\ [6pt]
\multirow{2}{*}{20 steps} 
& MAE & 0.912 & 5.99 \\ 
& RMAE & 0.413 \% & 2.71 \% \\ \hline
\end{tabular}
\end{table}

\subsection{Efficiency}\

Since the proposed method enables one-step generation, it achieves substantially higher computational efficiency than the finite element method. For a fair comparison, both model sampling and FEM simulations are performed on the same CPU platform. In each case, all 20 CPU cores are used for parallel execution, and the wallclock time is recorded for comparison. For each test, we perform 20 model inferences or 20 FEM simulations and report the average time, so as to ensure the comparison is as fair as possible. A state-of-the-art commercial solver, ABAQUS, is used for our finite element simulations. In addition, we also run the proposed model on a consumer-grade GPU, rather than a data-center-grade GPU, and measure the corresponding inference time for comparison. The results are summarized in Table \ref{tab:efficiency}. The speedup factor is defined as the time consumed by FEM divided by the time consumed by the proposed model. The results show that, for one-step generation, the proposed model achieves significant speedups of $6.61\times$ and $127\times$ over FEM on the CPU and GPU, respectively. Even when multi-step sampling is used to improve accuracy, the proposed model remains considerably faster than FEM because the required number of sampling steps is still small. 

Our framework also substantially outperforms existing approaches. As reported in \cite{kota2026hybrid}, even for a hyperelastic problem, diffusion models require more than 100 sampling steps to achieve satisfactory results. In another work \cite{jadhav2023stressd} concerning a linear elastic and hyperelastic problem, the authors employed as many as 500 sampling steps to generate the results, resulting in a sampling time of nearly 25 seconds even on a GPU.

\begin{table}[h]
\centering
\caption{Efficiency comparison between the proposed framework and FEM.}
\label{tab:efficiency}
\renewcommand{\arraystretch}{1.2}
\setlength{\tabcolsep}{12pt}
\begin{tabular}{c c c c c c}
\hline
Hardware & Metrics & FEM & One-step & 5 steps & 20 steps\\ \hline
\multirow{2}{*}{CPU} 
& Wallclock time (sec) & 70.1 & 10.6 & 17.3 & 42.5 \\ 
& Speedup factor & -- & 6.61 & 4.05 & 1.65 \\ [6pt]
\multirow{2}{*}{GPU} 
& Wallclock time (sec) & -- & 0.550 & 0.880 & 2.15 \\ 
& Speedup factor & -- & 127 & 79.7 & 32.6 \\ \hline
\end{tabular}
\end{table}

The runtime for one-step sampling is not approximately one-fifth of that required for 5 steps sampling. This is mainly because, regardless of the number of sampling steps, our VAE must be invoked 20 times, i.e., equal to the number of loading-unloading steps, to sequentially perform encoding and decoding. As a result, the computational cost of the VAE dominates the overall runtime in the one-step sampling setting. Furthermore, before the finite element simulation, our preprocessing code was executed in ABAQUS CAE and required an additional 10 seconds, most of which was spent on mesh generation. This computational cost is not included in Table \ref{tab:efficiency}. However, it should be noted that preprocessing and mesh generation consume non-negligible time costs when using the finite element method.




\section{Conclusion}\label{sec:Conclusion}

Despite the recent success of generative artificial intelligence and its growing attention in the engineering community, its application to computational mechanics still faces several limitations. First, diffusion models, which have been extensively studied, usually require hundreds of sampling steps, making their efficiency advantage over finite element analysis less pronounced. In contrast, flow matching remains underexplored in simulation-related applications. Second, existing GenAI-based simulation works are limited to linear elastic and hyperelastic problems, leaving path-dependent constitutive behaviors, such as plasticity, insufficiently investigated. Third, most existing flow matching models rely on Gaussian source distributions, while the potential benefits of non-Gaussian source distributions for improving model performance remain unclear.

In this work, a one-step generation flow matching framework with a spatiotemporal DiT backbone is proposed, which can efficiently generate plastic von Mises stress fields with random geometry and loading-unloading path. To achieve this approach, we designed a stochastic prior-information-based distribution as the source of flow matching to significantly mitigate crossings among conditional transport paths during training. We proposed token-level loading embedding method to improve the backbone's performance in plastic constitutive learning. Moreover, we designed a loss function for scientific VAE fine-tuning and introduced two auxiliary networks to improve our framework.

The experimental results demonstrate that, even with a limited training dataset size, our framework can accurately generate path-dependent stress fields, with both the overall stress magnitude and the spatial stress distribution effectively captured by the model. The proposed model also exhibits remarkably high efficiency. In the one-step generation setting, even on a CPU, our model provides more than a 6-fold speedup compared with finite element analysis. On a consumer-grade GPU, the model requires only 0.55 s to generate 20 frames of high-resolution stress fields with a resolution of $256 \times 256$. This research provides a new perspective for the application of GenAI in computational mechanics.

In future work, this study could be extended in several directions. First, the proposed method can be generalized to heterogeneous plastic materials and even fluid mechanics, thereby broadening the applicability of this one-step generative framework. Second, more efficient dimensionality-reduction techniques could be explored as well as improved representation strategies to facilitate the industrial-scale application of this approach. Finally, physics-informed extensions of the proposed framework could be devised, with the aim of reducing the amount of required training data without compromising model performance.


\section*{CRediT authorship contribution statement}

\textbf{Yijing Zhou:} Conceptualization, Methodology, Software, Validation,
Formal analysis, Investigation, Writing~-- original draft, Writing~-- review \& editing.
\textbf{Jasmin Jelovica:} Conceptualization, Supervision,
Writing~-- review \& editing, Funding acquisition.

\section*{Authors' note on AI Tools}

During the preparation of this work the authors used ChatGPT in order to enhance the readability. After using this tool/service, the authors reviewed and edited the content as needed and take full responsibility for the content of the publication.

\section*{Declaration of competing interest}

The authors declare that they have no known competing financial interests or personal relationships that could have appeared to influence the work reported in this paper.

\section*{Code availability}

Upon acceptance of the manuscript, the code associated with this research will be made publicly available on GitHub.

\section*{Acknowledgements}

This research was financially supported by the Natural Sciences and Engineering Research Council of Canada (NSERC) through Discovery Grant RGPIN-2025-04421 and Alliance Advantage Grant ALLRP 607677-25. Additional financial support was provided by Seaspan Shpyards and through the Four Year Doctoral Fellowship (4YF) from The University of British Columbia. Computational support was provided by Advanced Research Computing at The University of British Columbia and the Digital Research Alliance of Canada through access to research computing resources and services.

\appendix

\setcounter{figure}{0}

\section{Proof of Lemma 1}\label{app:proof}

\begin{proof}
Assume that the line segments $\overline{x_i y_i}$
and $\overline{x_j y_j}$ intersect at a common point $p$.

By the definition of a line segment, the collinearity of $p$ implies that the
total distance of each segment can be decomposed additively:
\begin{equation}
d(x_i,y_i) = d(x_i,p) + d(p,y_i),
\end{equation}
\begin{equation}
d(x_j,y_j) = d(x_j,p) + d(p,y_j).
\end{equation}

Next, we invoke the triangle inequality. For the triplets $(x_i,p,y_j)$ and
$(x_j,p,y_i)$, the following inequalities hold:
\begin{equation}
d(x_i,y_j) \le d(x_i,p) + d(p,y_j),
\end{equation}
\begin{equation}
d(x_j,y_i) \le d(x_j,p) + d(p,y_i).
\end{equation}

Summing the two inequalities yields
\begin{equation}
d(x_i,y_j) + d(x_j,y_i)
\le
d(x_i,p) + d(p,y_j) + d(x_j,p) + d(p,y_i).
\end{equation}

By rearranging the terms on the right-hand side and substituting the additive
distance expressions established for the intersecting segments, we obtain
\begin{align}
d(x_i,y_j) + d(x_j,y_i)
&\le
\left[d(x_i,p) + d(p,y_i)\right]
+
\left[d(x_j,p) + d(p,y_j)\right] \\
&=
d(x_i,y_i) + d(x_j,y_j).
\end{align}

However, summing the prescribed constraints
$d(x_i,y_i) < d(x_i,y_j)$ and $d(x_j,y_j) < d(x_j,y_i)$ gives the strict
inequality
\begin{equation}
d(x_i,y_i) + d(x_j,y_j)
<
d(x_i,y_j) + d(x_j,y_i).
\end{equation}

This is a contradiction, since the scalar quantity
$d(x_i,y_j) + d(x_j,y_i)$ cannot be simultaneously less than or equal to, and
strictly greater than, the scalar quantity
$d(x_i,y_i) + d(x_j,y_j)$.

Consequently, the initial assumption that such a point $p$ exists must be false.
Therefore, the line segments $\overline{x_i y_i}$ and $\overline{x_j y_j}$ do
not intersect.
\end{proof}

\section{Complete samples}\label{app:samples}

\begin{figure}[htp]
    \centering
    \includegraphics[width=\linewidth]{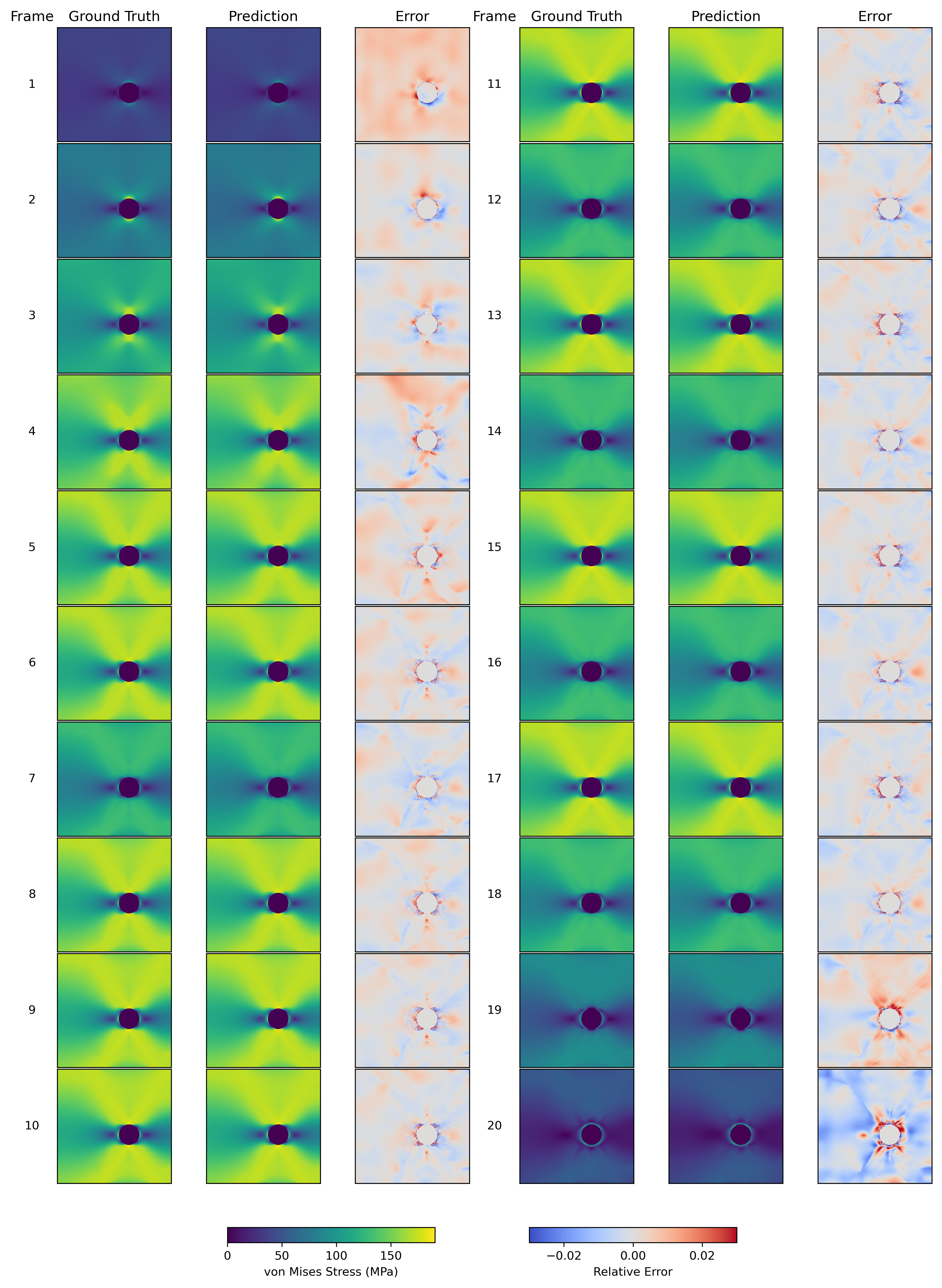}
    \caption{One-step generated sample of single hole case}
    \label{fig:Sample_1hole_1step}
\end{figure}

\begin{figure}[htp]
    \centering
    \includegraphics[width=\linewidth]{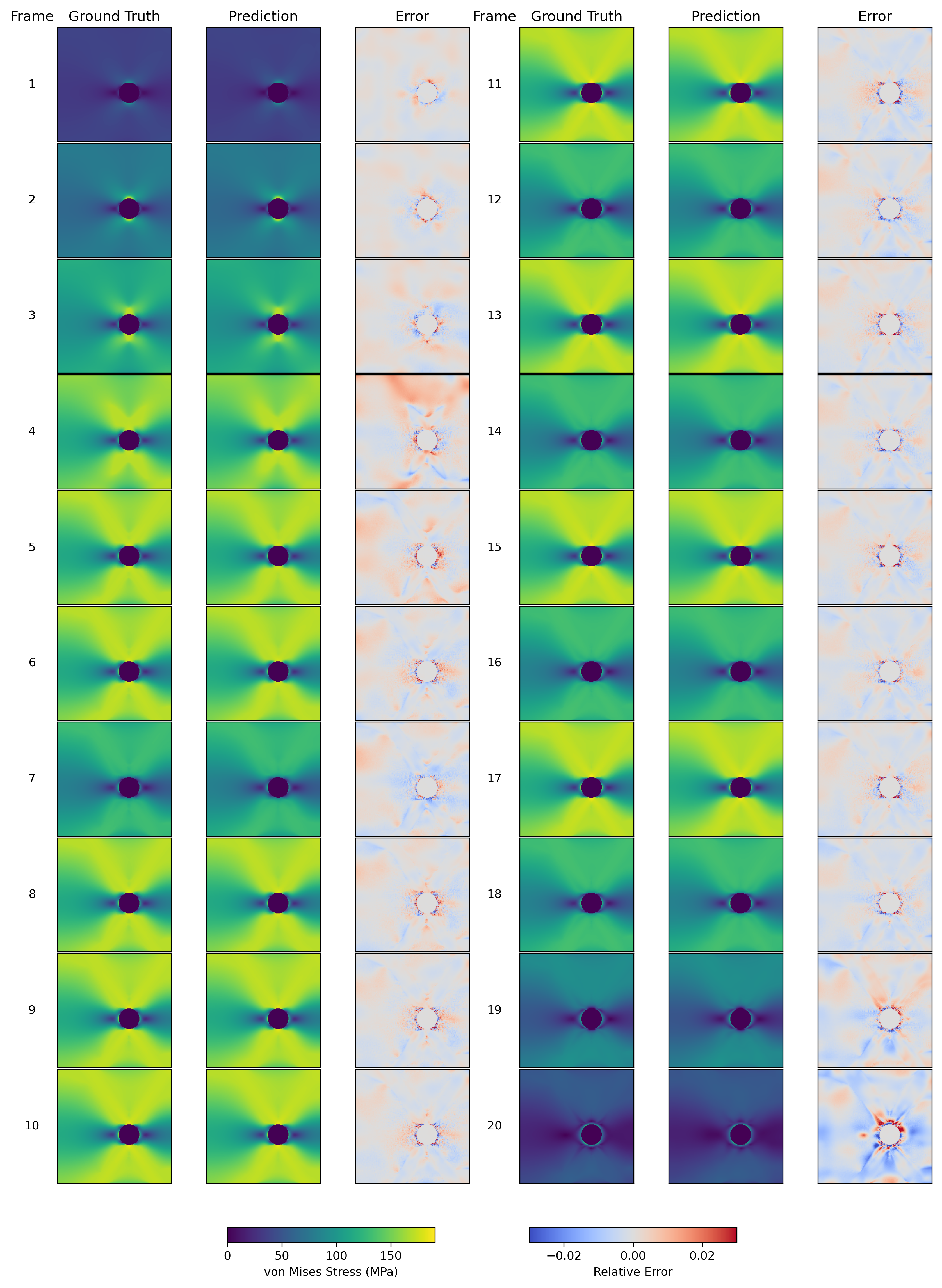}
    \caption{20 steps generated sample of single hole case}
    \label{fig:Sample_1hole_20steps}
\end{figure}

\begin{figure}[htp]
    \centering
    \includegraphics[width=\linewidth]{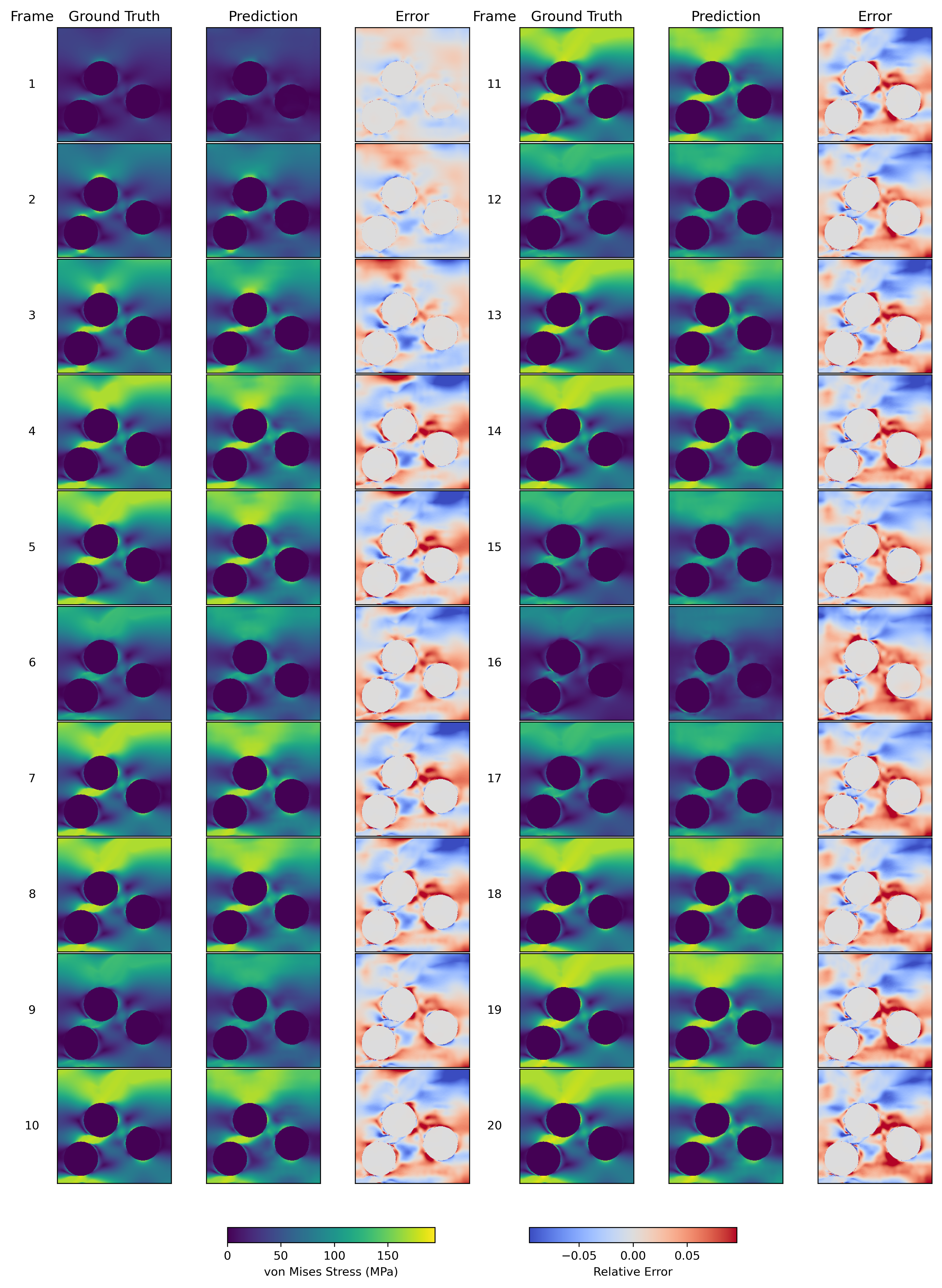}
    \caption{One-step generated sample of 3 holes case}
    \label{fig:Sample_3holes_1step}
\end{figure}

\begin{figure}[htp]
    \centering
    \includegraphics[width=\linewidth]{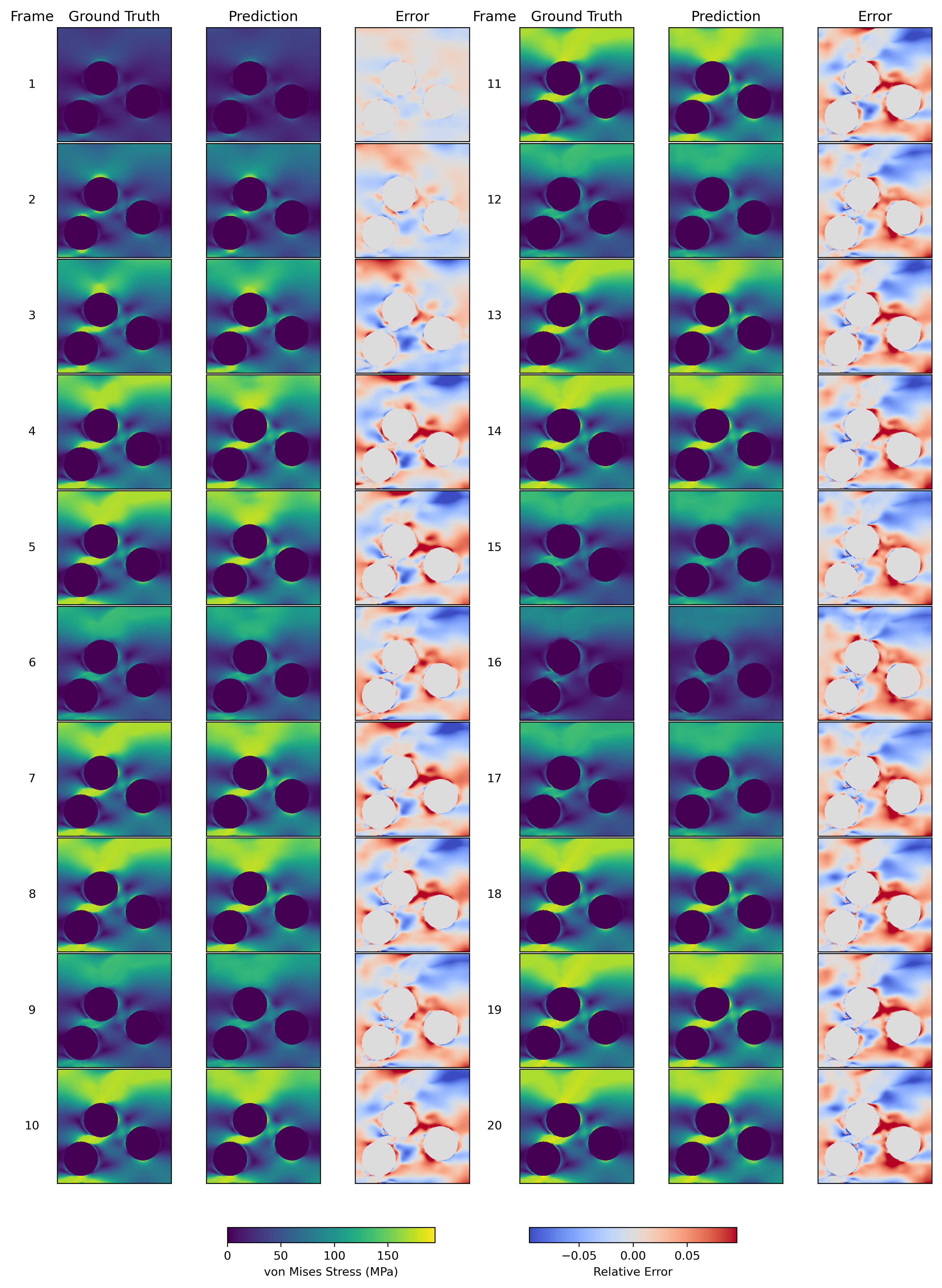}
    \caption{20 steps generated sample of 3 holes case}
    \label{fig:Sample_3holes_20steps}
\end{figure}

\begin{figure}[htp]
    \centering
    \includegraphics[width=\linewidth]{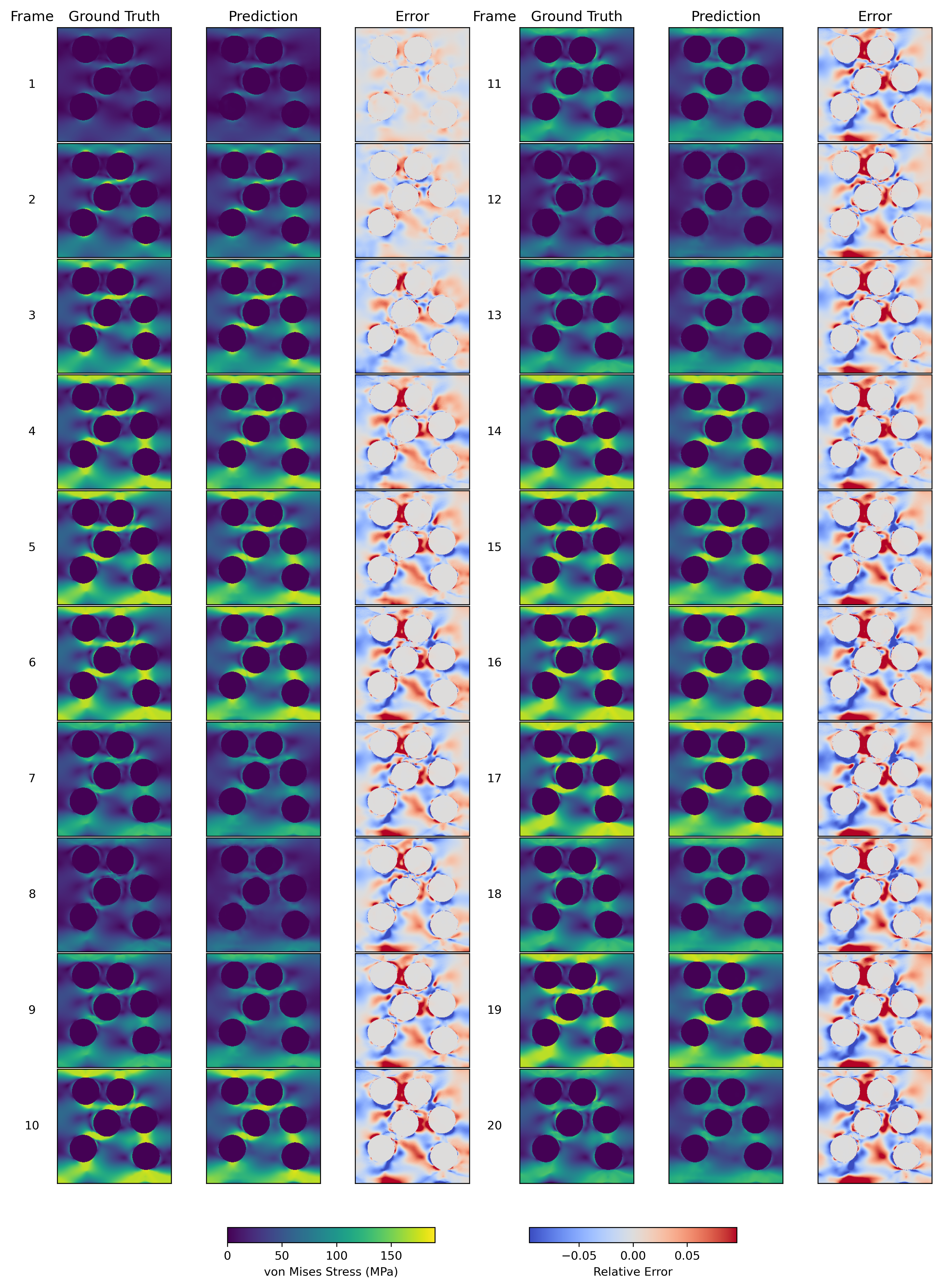}
    \caption{One-step generated sample of 6 holes case}
    \label{fig:Sample_6holes_1step}
\end{figure}

\begin{figure}[htp]
    \centering
    \includegraphics[width=\linewidth]{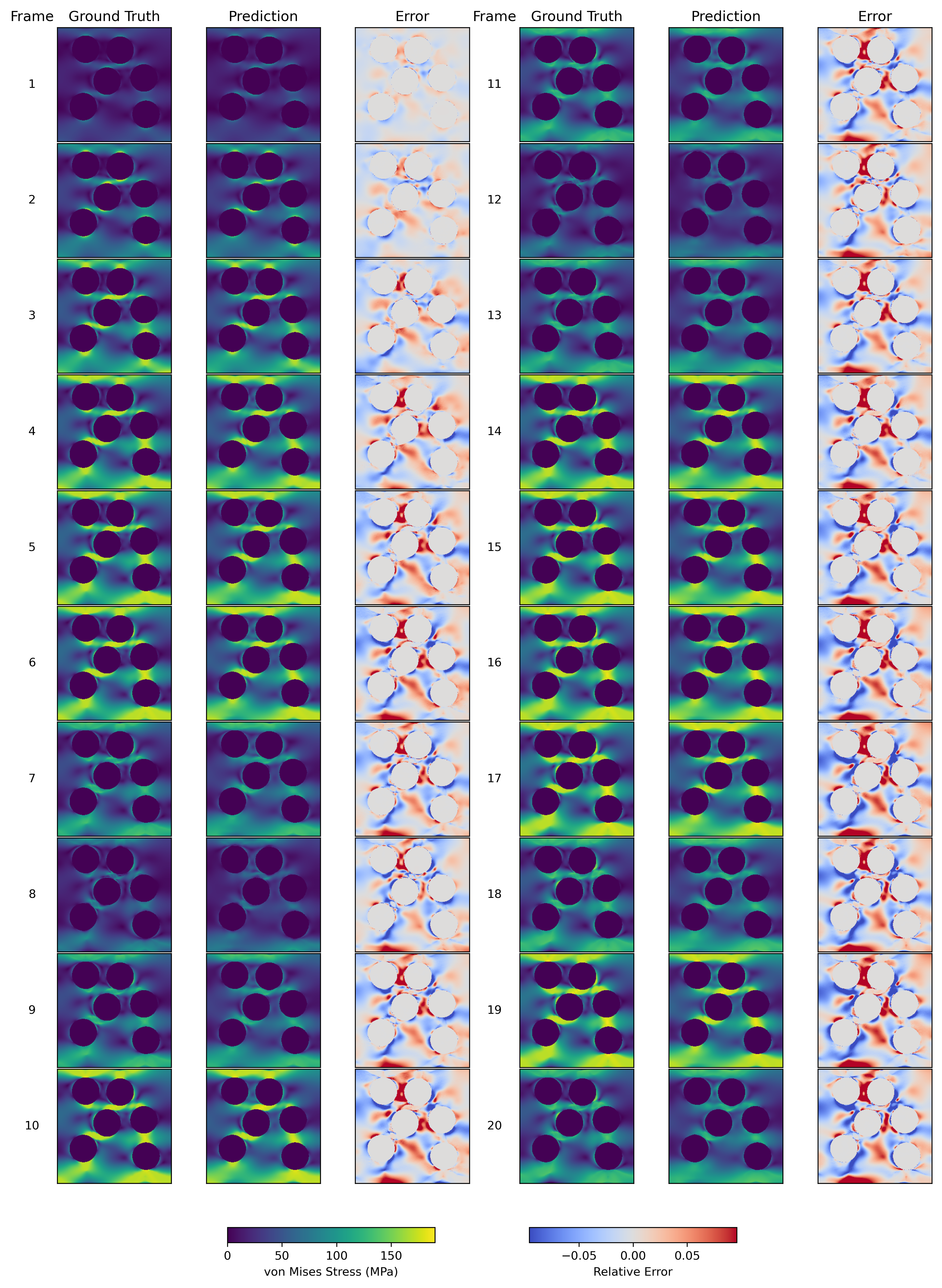}
    \caption{20 steps generated sample of 6 holes case}
    \label{fig:Sample_6holes_20steps}
\end{figure}

In this appendix, we provide frame-by-frame visualizations of the complete samples generated by our model. Figures \ref{fig:Sample_1hole_1step} and \ref{fig:Sample_1hole_20steps} present the single-hole case generated using one-step and 20 steps sampling, respectively. Figures \ref{fig:Sample_3holes_1step} and \ref{fig:Sample_3holes_20steps} present the 3 holes case generated using one-step and 20 steps sampling, respectively, while Figures \ref{fig:Sample_6holes_1step} and \ref{fig:Sample_6holes_20steps} show the corresponding results for the 6 holes case.

{\small
\bibliography{ref}
}
\end{document}